\documentclass[conference]{IEEEtran}
\IEEEoverridecommandlockouts
\usepackage{cite}
\usepackage{amsmath,amssymb,amsfonts}
\usepackage{algorithmic}
\usepackage{graphicx}
\usepackage{textcomp}
\usepackage{subcaption}
\usepackage[table,xcdraw]{xcolor}
\usepackage{multirow}
\usepackage{hyperref}
\usepackage{placeins}
\def\BibTeX{{\rm B\kern-.05em{\sc i\kern-.025em b}\kern-.08em
    T\kern-.1667em\lower.7ex\hbox{E}\kern-.125emX}}
\begin{document}

\title{Online Targetless Radar-Camera Extrinsic Calibration Based on the Common Features of Radar and Camera
\\
}

\author{\IEEEauthorblockN{Lei Cheng}
\IEEEauthorblockA{\textit{Department of Electrical and Computer Engineering} \\
\textit{University of Arizona}\\
Tucson, AZ, USA \\
leicheng@arizona.edu}
\and
\IEEEauthorblockN{Siyang Cao}
\IEEEauthorblockA{\textit{Department of Electrical and Computer Engineering} \\
\textit{University of Arizona}\\
Tucson, AZ, USA \\
caos@arizona.edu}
}

\maketitle

\begin{abstract}
Sensor fusion is essential for autonomous driving and autonomous robots, and radar-camera fusion systems have gained popularity due to their complementary sensing capabilities. However, accurate calibration between these two sensors is crucial to ensure effective fusion and improve overall system performance. Calibration involves intrinsic and extrinsic calibration, with the latter being particularly important for achieving accurate sensor fusion. Unfortunately, many target-based calibration methods require complex operating procedures and well-designed experimental conditions, posing challenges for researchers attempting to reproduce the results. To address this issue, we introduce a novel approach that leverages deep learning to extract a common feature from raw radar data (i.e., Range-Doppler-Angle data) and camera images. Instead of explicitly representing these common features, our method implicitly utilizes these common features to match identical objects from both data sources. Specifically, the extracted common feature serves as an example to demonstrate an online targetless calibration method between the radar and camera systems.  The estimation of the extrinsic transformation matrix is achieved through this feature-based approach. To enhance the accuracy and robustness of the calibration, we apply the RANSAC and Levenberg-Marquardt (LM) nonlinear optimization algorithm for deriving the matrix. Our experiments in the real world demonstrate the effectiveness and accuracy of our proposed method. 
\end{abstract}


\section{Introduction}
\begin{figure} 
	\centering
	\includegraphics[width=0.45\textwidth]{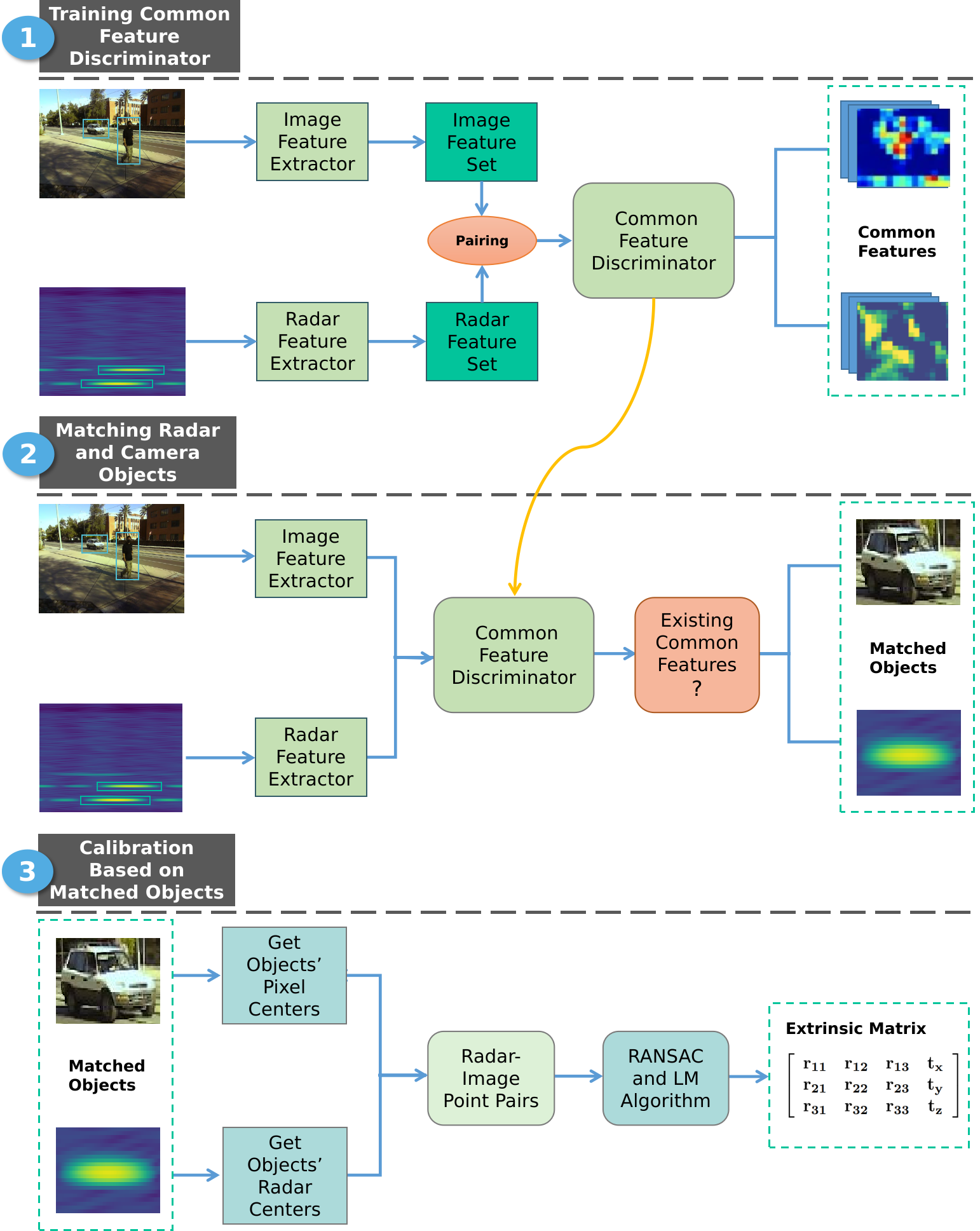}
	\caption{Framework for the proposed radar-camera online extrinsic co-calibration method. The pipeline showcases the sequential steps involved in calibrating the radar and camera sensors. The method first trains a deep-learning common feature discriminator to determine whether the detected objects in the radar and camera data share common features. Subsequently, the trained common feature discriminator is utilized to find matching objects in both radar and camera views based on the existence of common features. Finally, based on these matching objects, corresponding camera-radar point pairs are formed for calibration.}
	\label{0}
\end{figure}

Radar and camera are complementary sensing modalities widely used in applications such as autonomous driving and robotics. Radar provides accurate range, velocity, and angle information regardless of illumination and weather conditions, while the camera captures high-resolution visual information. Combining radar and camera improves perception and enables tasks like object recognition, detection, and tracking in dynamic environments. Precise calibration between radar and camera is crucial for accurate sensor fusion due to their differing sensing principles, especially for determining their relative pose (extrinsic calibration).
Extrinsic calibration estimates the spatial transformation of the sensor coordinates to other sensors or unified reference frames\cite{domhof2019extrinsic}. One way to achieve it is through target-based methods that use specially designed calibration targets such as checkerboards or corner reflectors. However, these methods are highly sensitive to the accuracy and availability of the targets and necessitate the preparation of specific calibration scenes. Furthermore, they are unable to handle runtime decalibrations that frequently occur in real-world applications\cite{scholler2019targetless}, leading to potential inaccuracies in the calibration results. In contrast, targetless methods circumvent the need for external targets and instead focus on detecting and matching natural features present in the scene that are observable by both the radar and camera sensors. By leveraging these common features, targetless methods eliminate the reliance on specific calibration targets, making them more flexible and adaptable in diverse real-world environments. While targetless methods do not require many manual steps and a well-controlled experimental environment, they still face challenges such as requiring good initial calibration and relying on high-quality natural features and their accurate extraction. One promising approach to address these challenges is using deep learning to extract useful features from raw radar data and explore the relationship between radar and image features for estimating the extrinsic transformation matrix, as shown in Fig. \ref{0}. Unlike traditional target-based methods, the proposed approach not only reduces the complexity and setup requirements but also enables the system to perform online recalibration, mitigating the potential deterioration of the extrinsic matrix over time. This capability is particularly crucial in dynamic environments where factors such as vibrations, temperature variations, and general wear and tear can affect the spatial alignment between the radar and camera sensors.
The contributions of this work are highlighted as follows: 
\begin{enumerate}
  \item  Developing a deep learning model that learns useful features from raw radar data and explores the relationship between radar and image features to obtain common features.
  \item  The first known method that leverages the common features of radar and camera to implement an online targetless calibration approach and addresses the challenges of traditional calibration methods, such as the need for specific calibration targets and manual steps, making the proposed approach more practical and effective for real-world applications.
\end{enumerate}
The rest of the paper is organized as follows. In Section II, we review related work on calibration methods for sensor fusion. In Section III, we describe our proposed online targetless calibration approach in detail. In Section IV, we present and discuss our experimental results. Finally, we conclude the paper in Section V and discuss future research directions.

\section{Related Work}
Extrinsic calibration methods for radar-camera systems can be categorized into target-based and targetless approaches. Target-based methods utilize calibration targets but may be infeasible and introduce errors in real-world scenarios. Targetless methods extract features from natural scenes, making them more suitable for practical scenarios. Our research focuses on developing an online targetless calibration method that leverages deep learning techniques to extract common features.
\subsection{Targetless Extrinsic Calibration Method}
Wisec et al. \cite{wise2021continuous} perform targetless extrinsic calibration of 3D radar and camera by relying on velocity information provided by the 3D radar, instead of attempting to localize and track specific targets. This method considers a continuous-time batch nonlinear optimization problem with its cost function based on radar velocity measurements and camera pose measurements from specific motions. The Levenberg-Marquardt (LM) algorithm is used to solve this problem. But over-reliance on radar velocity measurements makes it less robust to noisy radar measurements. Wisec et al. \cite{wise2022spatiotemporal} recently have extended their previous work by proposing an algorithm that utilizes radar ego-velocity estimates, unscaled camera pose measurements, and a continuous-time trajectory representation to achieve spatiotemporal calibration between radar and camera. Per{\v{s}}i{\'c} et al.\cite{pervsic2021online}  propose an online targetless multi-sensor calibration method based on the detection and tracking of moving objects. It employs the tracking-based decalibration detection and a single graph-based optimization. Nevertheless, this method is only limited to rotation calibration without considering translation. 
Sch{\"o}ller et al. \cite{scholler2019targetless} employed a deep learning approach, training coarse followed by fine convolutional neural networks (CNNs), to learn the rotational calibration matrix from camera images and radar point-cloud data. However, this approach still necessitates the manual measurement of the translation parameters and relies on the manual determination of ground truth calibration.
Heng et al. \cite{heng2020automatic} proposed a calibration method that leverages a highly precise 3D map constructed with 3D LiDAR sensors to register radar scans and achieve accurate calibration for the lidar-radar system. However, this approach is not applicable for camera-radar calibration, as cameras cannot generate highly precise 3D maps like LiDAR sensors.

\begin{figure*} 
	\centering
	\includegraphics[width=0.7\textwidth]{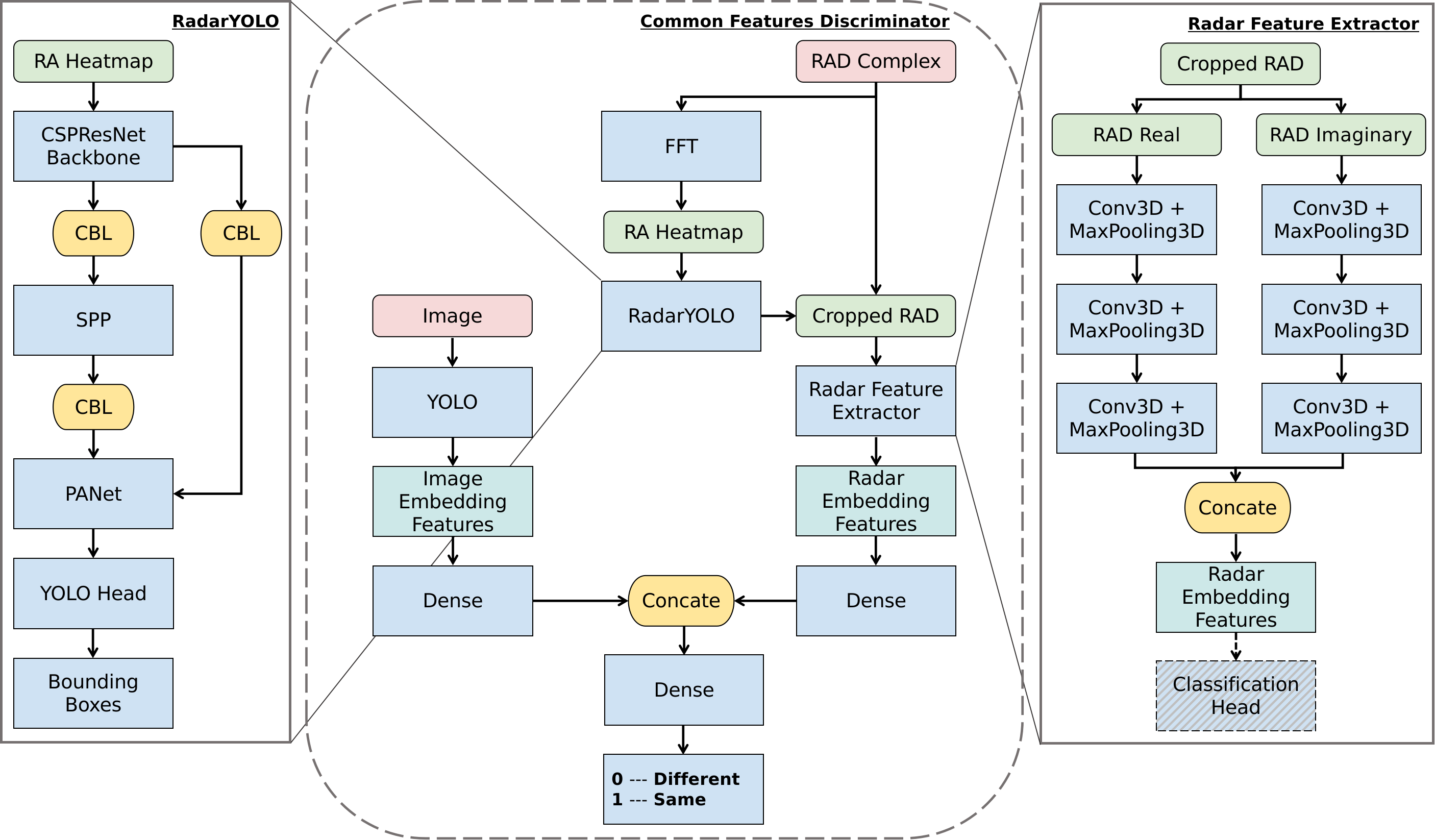}
	\caption{Architecture of the YOLO-based Common Feature Network. CSPResNet: Cross-Stage Partial ResNet.  CBL: Convolution3D + Batch Normalization + LeakyReLU. SPP: spatial pyramid pooling. PANet:  Path Aggregation Network \cite{bochkovskiy2020yolov4}.}
	\label{2}
\end{figure*}
\subsection{Deep Learning for Radar}
Targetless calibration can be addressed by leveraging either the motion of objects or the feature correspondences in the environment. Currently, motion-based methods are more commonly used in radar-involved calibration, as feature-based methods heavily rely on detailed structural information that radar sensors may not provide \cite{pervsic2021online,scholler2019targetless,li2023globally}. However, recent advancements in deep learning and neural networks have shown promise in extracting meaningful features directly from radar data \cite{bhattacharya2020deep,patel2019deep}. Deep neural networks can learn complex representations and extract discriminative features from radar signals \cite{wang2019study}, enabling feature-based calibration. 
Bhattacharya et al. \cite{bhattacharya2020deep} proposed a CNN model named "RadarNet" that consists of a convolutional block, three inception modules, and pooling layers for binary classification tasks based on radar spectrograms. Patel et al. \cite{patel2019deep} proposed another simple CNN architecture with 3 convolutional and 2 fully-connected layers for multi-class classification tasks using radar range-velocity-azimuth spectra. Wang et al. \cite{wang2021rodnet} introduced RODNet, a radar object detection model that utilizes a 3D CNN architecture. This model is cross-supervised by a novel 3D localization of detected objects using a camera-radar fusion (CRF) strategy and takes a snippet of RF images as the input to do radar object detection. Zhang et al. \cite{zhang2021raddet} proposed another model called RadarResNet, which utilizes a backbone network based on ResNet architecture. This model can perform 3D detections in the Range-Azimuth-Doppler map. In addition, the success of the YOLO (You Only Look Once) \cite{bochkovskiy2020yolov4} algorithm in image detection has also sparked interest in applying YOLO-based approaches to process radar data. Song et al.\cite{song2022ms} introduced MS-YOLO, a modified version of the original YOLO algorithm, with the aim of enhancing detection accuracy by incorporating and extracting millimeter-wave radar and visual features. Huang et al.\cite{huang2022yolo} introduced YOLO-ORE, which combines YOLO with a radar image generation module, to reduce overlap errors and misclassification errors of YOLO.
Due to the proven effectiveness of YOLO in various object detection tasks and its ability to directly learn intricate patterns and features from radar data, we have opted to adopt the YOLO approach for finding common features in this work.

\section{Proposed Method}
\subsection{Problem Formulation}\label{AA}
Radar-camera calibration involves solving for a transformation matrix that establishes a correspondence between a point in the image pixel coordinate system (PCS) and another point in the radar coordinate system (RCS). This transformation matrix comprises the intrinsic matrix and the extrinsic matrix, which are obtained through intrinsic calibration and extrinsic calibration, respectively. In radar-camera extrinsic calibration, the goal is to determine the extrinsic matrix using the known camera intrinsic parameters and a set of N correspondences between the points in RCS and their corresponding points in PCS. 
Assuming that a point in the RCS is denoted as $P_r$, and its corresponding point in the PCS is denoted as $P_p$, the transformation between them can be represented as:
\begin{equation}\label{eq1}
s\ P_p  = K\ [R|T]\ P_r =K Q P_r
\end{equation}
where $K$ is the intrinsic matrix and $Q$ is the extrinsic matrix. The extrinsic matrix is composed of the rotation matrix $R$ and the translation matrix $T$. Given that more than 3 point correspondences are used, the calibration problem transitions into minimizing the reprojection error, which represents the residual errors between the projected pixel points of the radar points and their corresponding ground truth pixel points. This problem is commonly known as the Perspective-n-Point (PnP) problem. One type of reprojection error, known as the root-mean-square of reprojection errors, can be defined as:
\begin{equation}\label{eq9}
\begin{split}
Err_{rep}
&=\sqrt{\frac{1}{N}\sum_{i=1}^{N}\left\| P^{i}_{p\_gt}-P^{i}_{p} \right\|_{2}^{2}}\\
&=\sqrt{\frac{1}{N}\sum_{i=1}^{N}\left\| P^{i}_{p\_gt}-s^{-1}KQP^{i}_r \right\|_{2}^{2}}
\end{split}
\end{equation}
where $N$ is the number of point correspondences, and $P_{p\_gt}$ is the ground truth pixel point directly obtained from the image.

\subsection{Point Correspondences Matching Based On Common Features}
Based on the previous discussion, the key to solving the extrinsic matrix, which relates to the radar and camera coordinate systems, is to find a sufficient number of point correspondences between the radar and camera views. Traditionally, researchers have relied on the use of distinctive calibration targets, such as corner reflectors, to enhance the detectability of objects in both radar and camera images. These targets serve as reference points that facilitate the establishment of accurate point correspondences between the two modalities. However, for targetless calibration methods, we do not use specific calibration targets and do not have prior knowledge of the objects in the radar and camera views. Fortunately, based on the shared characteristics observed in radar and camera detections corresponding to the same objects, we can match these objects and further align their centers, resulting in point correspondences. The common features extracted from both radar and camera detections provide valuable information for identifying and associating objects across the two modalities. These features can include reflectivity, texture, size, and motion patterns. By analyzing these features, we can identify objects that exhibit similar patterns or characteristics in both radar and camera data.

Due to inherent limitations such as low resolution and high noise in radar data, extracting features using conventional algorithms can be challenging. However, deep learning has shown remarkable capabilities in feature extraction, making it well-suited for tackling these challenges. Given the success of the YOLO model in image feature extraction and our objective of investigating shared features between radar and camera data, we have opted to employ YOLO-based methods for extracting common features. 
By adapting the YOLO model to radar data, we can leverage its powerful feature extraction capabilities to identify and extract pertinent radar features. This empowers us to detect objects in radar frames. Simultaneously, the YOLO model's ability to detect and recognize objects in camera images allows us to identify common features between radar and camera data. Furthermore, based on the extracted radar and image features, we construct another deep neural network that specifically focuses on discovering the shared characteristics and matched features between the two modalities. This network is designed to learn the intricate relationships and patterns that exist between the extracted features, allowing us to identify and align the corresponding features from radar and camera data. Ultimately, we can effectively explore and exploit the common features shared by radar and camera data. 

Fig. \ref{2} visually presents the structure of the YOLO-based common feature network and highlights its key components and layers for comprehensive understanding.
The network comprises three main components: a YOLO-based radar detector, which takes fast Fourier transform (FFT) heatmaps generated from the raw radar data as input and outputs bounding boxes containing the detected targets; a 3D convolutional neural network (CNN) radar feature extractor, which takes the bounding boxes obtained from the radar detector and crops the range-azimuth-doppler (RAD) data enclosed by these bounding boxes as input, and then determines the quality of the extracted features based on the accuracy of classification; and a fully connected common feature discriminator, which utilizes the image features extracted by YOLO and the radar features extracted by the radar detector and radar feature extractor to determine whether an object detected in an image and an object detected by the radar are the same entity.

\subsection{Radar-Camera Extrinsic Calibration Solution}
With the common features extracted through deep learning, we can perform object matching and correspondence identification. By comparing the features of objects detected in radar and camera views, we can determine which objects correspond to each other. Then, we can extract the centers of these objects by utilizing the spatial information encoded in the common features, which enables us to establish point correspondences between radar and image pixel coordinate systems.

Based on the obtained point correspondences, we can proceed with the estimation of the extrinsic calibration matrix. This is inherently a nonlinear problem for minimizing the reprojection error. The LM algorithm, which combines the advantages of steepest descent and Gauss-Newton methods, has been widely adopted for solving nonlinear least-squares problems\cite{marchand2015pose}, including our case. However, to ensure convergence to the globally optimal solution, the algorithm requires a good initial guess for the extrinsic matrix. In our approach, an initial estimate of the calibration matrix based on the RANSAC (Random Sample Consensus)\cite{fischler1981random} algorithm served as the initial guess for the subsequent LM minimization. RANSAC is an iterative method that aims to estimate parameters of a mathematical model using the smallest set of possible correspondences while effectively rejecting outliers. In the context of extrinsic calibration, RANSAC helps to address the challenges posed by noisy radar measurements and variations in the radar-camera point correspondences. 
After using RANSAC to filter out unsuitable correspondences and obtain a robust initial estimate of the calibration matrix, the remaining point correspondences are then used as input to solve the PnP problem. This two-step approach combines the strengths of RANSAC in handling outliers and the accuracy of the iterative LM optimization algorithm, resulting in an effective solution for extrinsic calibration in the presence of noisy radar measurements and variations in point correspondences.


\begin{figure*}[htbp]
    \centering
    \begin{subfigure}[t]{0.38\textwidth}
        \centering
        \includegraphics[width=\textwidth]{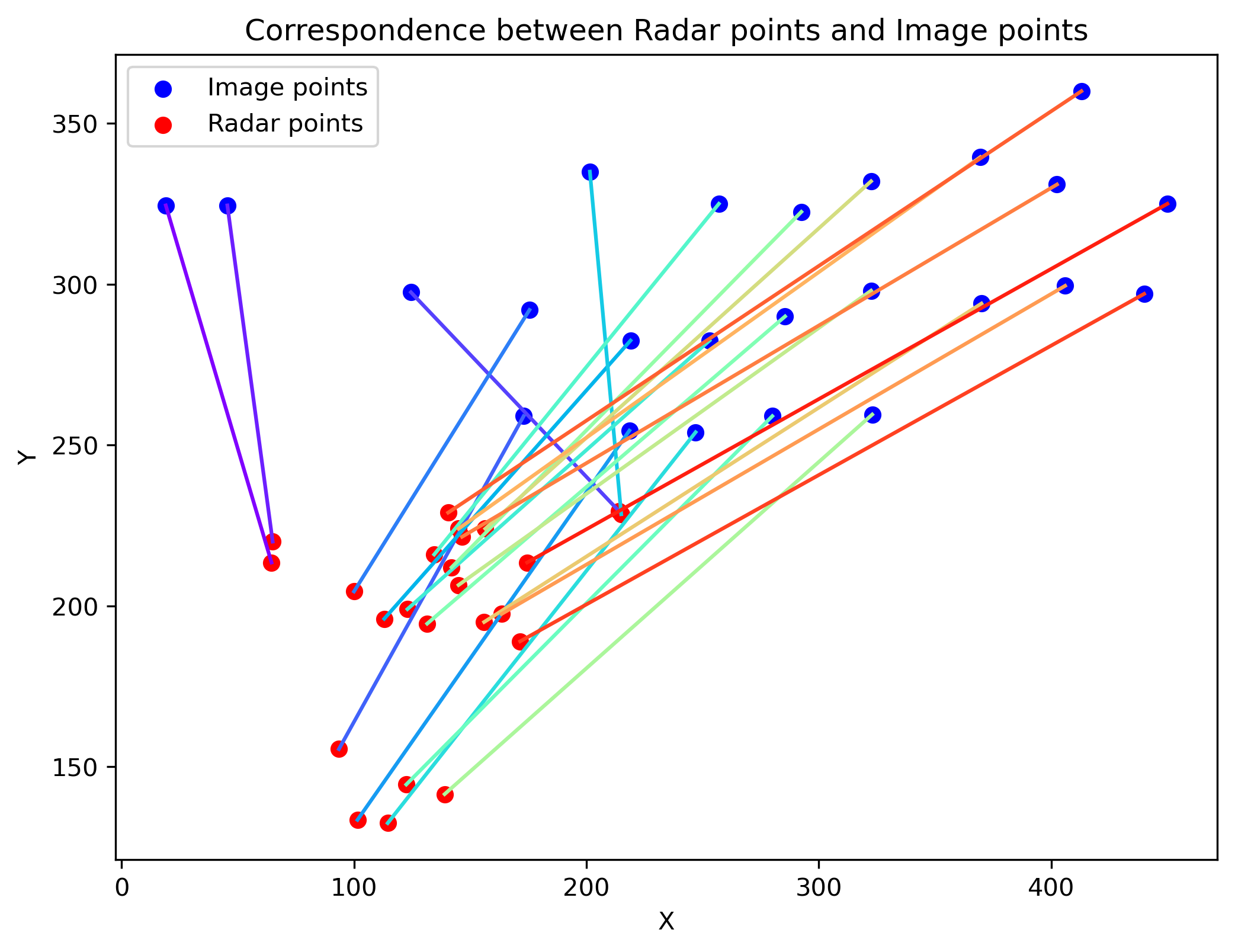}
        \caption{}
        \label{fig:org_corresp}
    \end{subfigure}\hspace{0.06\textwidth}
    \begin{subfigure}[t]{0.38\textwidth}
        \centering
        \includegraphics[width=\textwidth]{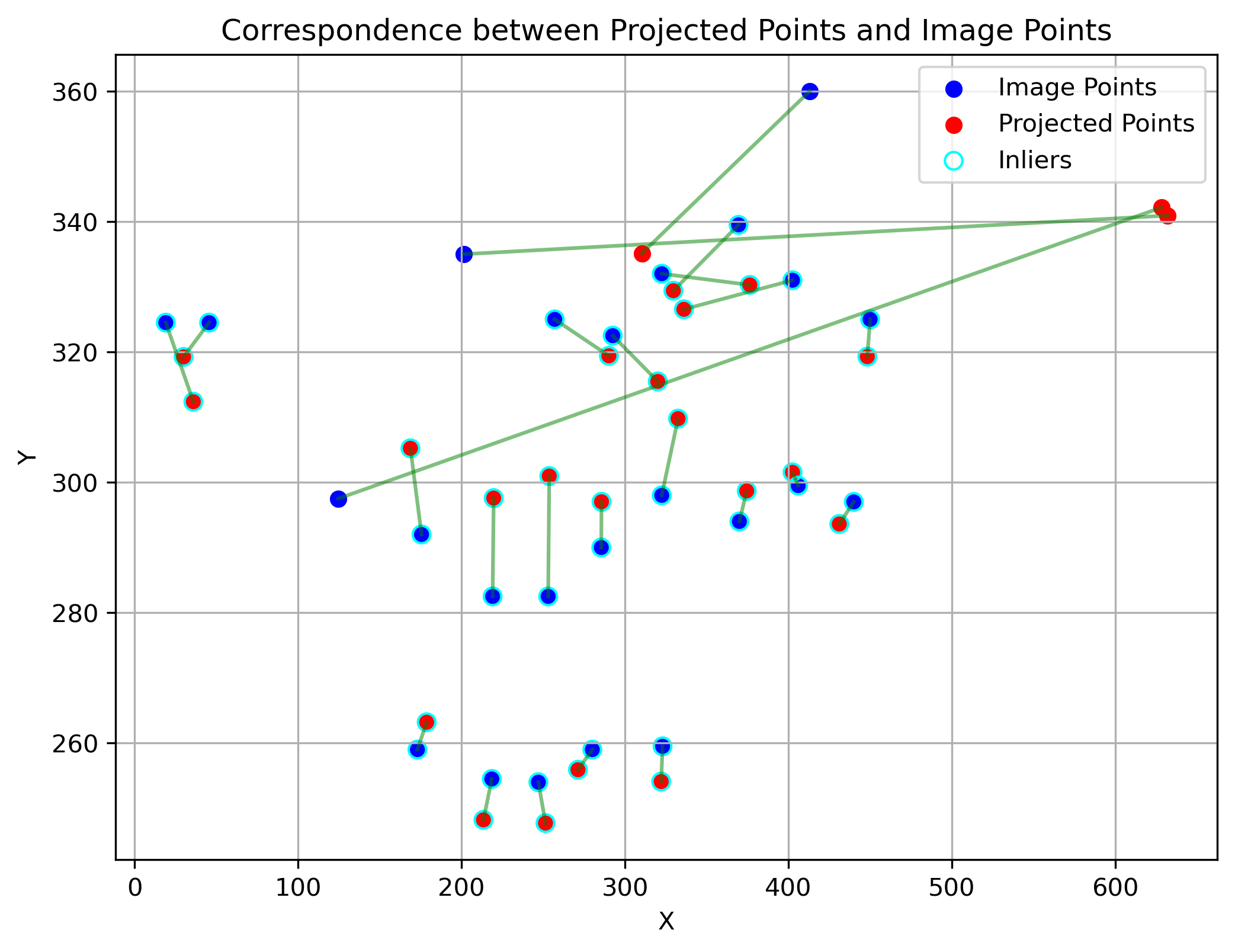}
        \caption{}
        \label{fig:transf_corresp}
    \end{subfigure}
    \caption{(a) The $24$ image-radar point correspondences obtained through a block-based sampling strategy for calibration. (b) The correspondence between the projected radar points (i.e., the radar points from (a) projected onto the image using the calibration matrix) and the image points, as well as the inliers used for calibration.}
    \label{fig:org_transf_corresp}
\end{figure*}

\begin{figure*}[htbp]
    \centering
    \begin{subfigure}[t]{0.3\textwidth}
        \centering
        \includegraphics[width=\textwidth]{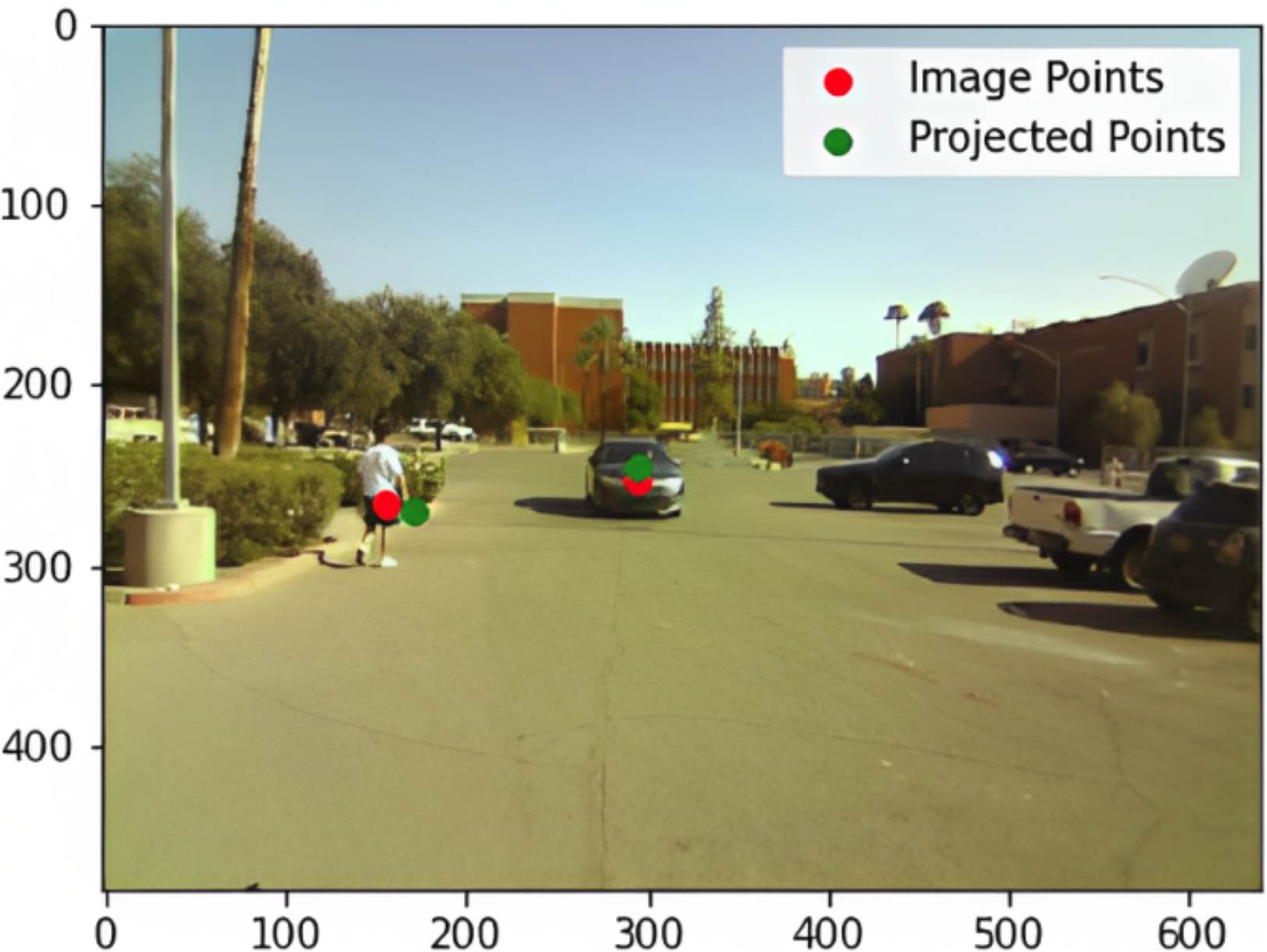}
        \caption{}
        \label{fig:image1}
    \end{subfigure}%
    \hfill
    \begin{subfigure}[t]{0.3\textwidth}
        \centering
        \includegraphics[width=\textwidth]{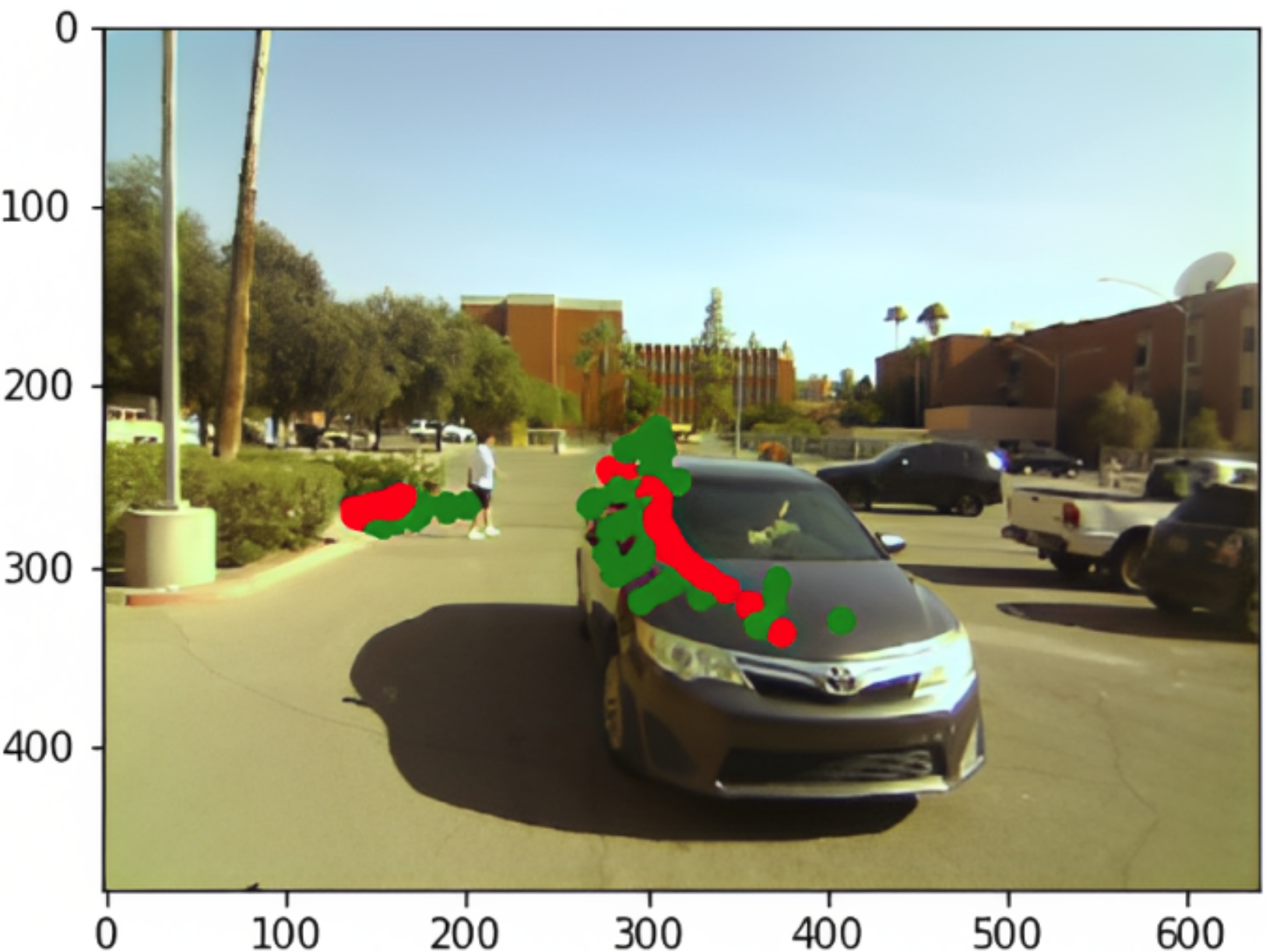}
        \caption{}
        \label{fig:image2}
    \end{subfigure}%
    \hfill
    \begin{subfigure}[t]{0.32\textwidth}
        \centering
        \includegraphics[width=\textwidth]{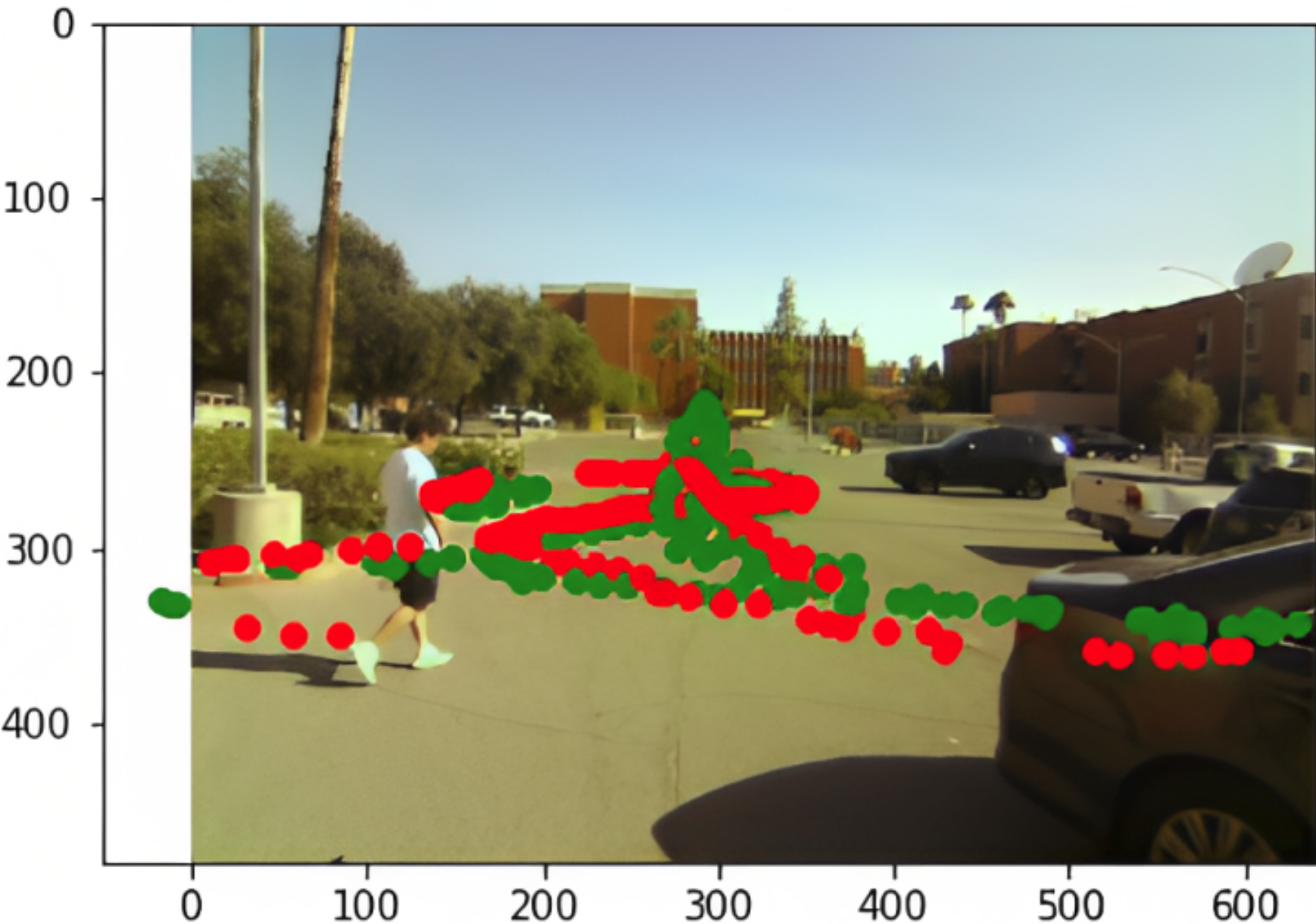}
        \caption{}
        \label{fig:image3}
    \end{subfigure}
    \caption{Projecting radar points onto the image using the obtained calibration matrix. (a) Projection of individual radar points onto the corresponding targets in the image, namely the car and the person. (b) Trajectories of the image points and the projected radar points corresponding to the movement of the two targets at frame 39. (c) Trajectories of the image points and the projected radar points corresponding to the movement of the two targets at frame 204.}
    \label{fig:side_by_side_images}
\end{figure*}
\begin{figure}[h!]
	\centering        \includegraphics[width=0.38\textwidth]{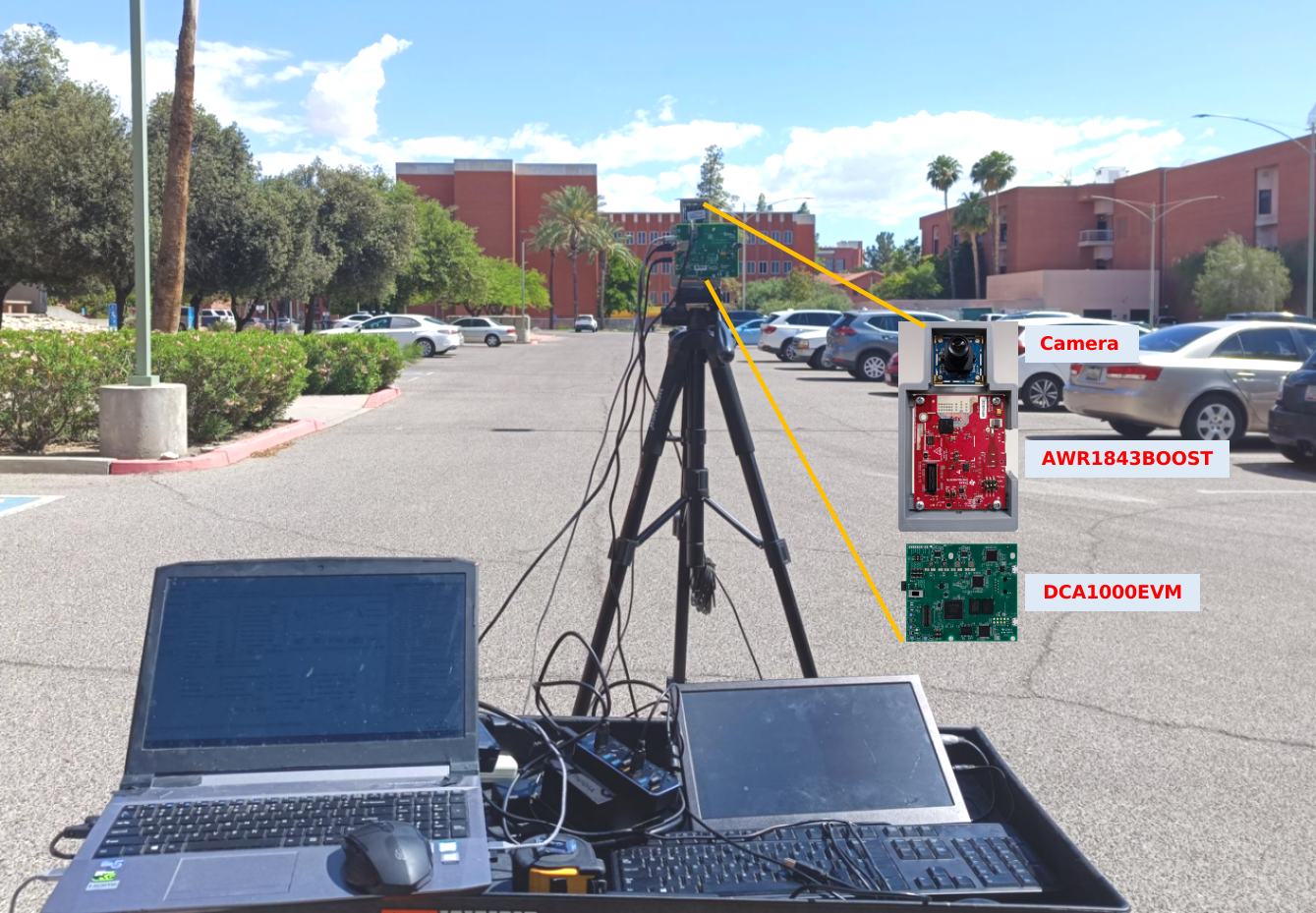}
	\caption{The experimental setup, where a laptop and an NVidia Jetson Xavier GPU establish a connection with the radar-camera system using Ethernet and USB cables to collect data.}
	\label{3}
\end{figure}
\section{Experiments And Results}
\subsection{Experimental Setup}
We use (i) a TI AWR1843BOOST mmWave radar transceiver with antenna channels in azimuth and elevation to allow for 3D sensing, and (ii) a USB8MP02G monocular HD digital camera as our two sensing modalities. To capture radar raw data, we employed a TI DCA1000 evaluation module (EVM). These sensors were placed on a static tripod mounted using a 3D printed frame. An NVidia Jetson Xavier running Ubuntu 18 and ROS Melodic is used to launch YOLO for capturing images from the camera. Additionally, a Windows laptop was used to collect radar raw data by running the mmWave studio GUI tool. The experimental setup is illustrated in Fig. \ref{3}.

\begin{table}[htbp]
    \centering
    \caption{Reprojection errors for the estimated calibration matrix}
    \renewcommand{\arraystretch}{1.2}
    \rowcolors{2}{gray!15}{white}
    \begin{tabular}{|c|c|c|}
        \hline
        \rowcolor{gray!50}
        \textbf{Metric} & \textbf{All Points} & \textbf{Inliers} \\
        \hline
        \cline{1-3} 
        Mean Absolute Reprojection Error & 59.89 & 18.83 \\
        Root Mean Squared Reprojection Error & 98.48 & 17.75 \\
        \cline{1-3} 
        \hline
        Number of points & 24 & 21 \\
        \hline
    \end{tabular}
    \label{tab:calibration_metrics}
\end{table}
\subsection{Results and Discussion}
The experiment was conducted in a parking lot with a human subject and a car. They moved freely within the parking lot while data was collected. Since we performed online calibration, we used the data collected during the first minute of the experiment (with a frame rate of 30 frames per second for both the camera and radar, resulting in 1800 frames) to calibrate and obtain the calibration matrix. Then, we used this calibration matrix to project the radar points collected after the first minute onto the image, allowing us to evaluate the accuracy of the calibration matrix. 

During the calibration process, we employed a block-based approach to partition the image into many small blocks, each measuring $20\times20$ pixels. The image-radar point correspondences collected during the first minute were assigned to these blocks based on their spatial location within the image. Specifically, we selected one block at intervals of one block and identified the point correspondence closest to the center of that block. This procedure yielded a set of point correspondences that were utilized to solve the calibration matrix. By incorporating this block-based sampling strategy, we ensured that the calibration matrix was computed using a representative set of point correspondences that captured the spatial distribution, enabling us to obtain accurate calibration results.

In this experiment, a total of $24$ point correspondences(as shown in Fig. \ref{fig:org_corresp}) were obtained for calibration. This was due to the relatively large size of the blocks used for partitioning, many points fell within the same block, resulting in multiple point correspondences assigned to the same region. 

Table \ref{tab:calibration_metrics} presents the evaluation metrics for assessing the accuracy of the calibration matrix, including the Mean Absolute Reprojection Error (MARE) and the Root Mean Squared Reprojection Error (RMSRE). The MARE represents the average absolute difference between the projected radar points and their corresponding image points, while the RMSRE measures the overall deviation between the projected and actual image points. For all points, the MARE was measured as $59.89$ pixels, and the RMSRE was found to be $98.48$ pixels. Although this may seem like a significant difference, it is primarily attributed to the presence of outliers (as shown in Fig. \ref{fig:transf_corresp}). When considering only the inliers, these errors (less than $20$ pixels) fall within an acceptable range. Additionally, the evaluation encompasses a total of $24$ points, with $21$ identified as inliers. This indicates that the calibration process effectively handles the noisy radar measurements in the radar-camera correspondences, resulting in improved accuracy and robustness.

The evaluation of our calibration performance is further supported by the results of projecting radar points onto the image using the obtained calibration matrix, as shown in Fig. \ref{fig:side_by_side_images}. From Fig. \ref{fig:image1}, it is evident that the image points and the projected radar points are very close to each other, with the projected radar points predominantly falling within the region occupied by the targets (i.e., the car and the person) in the image. Fig. \ref{fig:image2} and Fig. \ref{fig:image3} display the trajectories of the image points and the projected radar points corresponding to the movement of the two targets. It can be observed that the radar and image trajectories exhibit good alignment, particularly considering the high variability of the radar points. These visual results highlight the practicality and effectiveness of the proposed calibration method, even in the face of inherent fluctuations in the radar measurements.




\section{Conclusions}\label{SCM}
We present an online targetless radar-camera extrinsic calibration method based on the common features of radar and camera. The proposed approach leverages deep learning techniques to extract common features from raw radar data and camera images, enabling the estimation of the extrinsic transformation matrix without the need for specific calibration targets. Experiments validate the effectiveness of the proposed method, demonstrating its potential for reliable and precise radar-camera extrinsic calibration in challenging real-world environments. This research contributes to the field of radar-camera sensor fusion and provides a foundation for further advancements in calibration techniques and expanding the application of the proposed method. Future work may focus on further enhancing the efficiency and adaptability of the calibration method and exploring visualizing and interpreting the common features between radar and camera data.  

\section*{Acknowledgment}
This project was funded by the National Institute for Transportation and Communities (NITC; grant number 1296), a U.S. DOT University Transportation Center, and also supported by the Tucson Department of Transportation.

\bibliographystyle{ieeetr}
%
{\small
\bibliography{references.bib}
}
\vspace{12pt}
\color{red}

\end{document}